# Estimating optical vegetation indices and biophysical variables for temperate forests with Sentinel-1 SAR data using machine learning techniques: A case study for Czechia

*Daniel Paluba [1], Bertrand Le Saux [2, 3], Přemysl Štych [1]*

[1] EO4Landscape Research Team, Department of Applied Geoinformatics and Cartography, Faculty of Science, Charles University, Prague, Czechia; contact: palubad@natur.cuni.cz, stych@natur.cuni.cz
[2] Φ-lab - Climate Action, Sustainability and Science Department (EOP-S) Department, Earth Observation Programmes Directorate, European Space Agency (ESA/ESRIN), Frascati, Italy
[3] AI4EARTH, Toulouse, France.

## Abstract

Current optical vegetation indices (VIs) for monitoring forest ecosystems are well established and widely used in various applications, but can be limited by atmospheric effects such as clouds. In contrast, synthetic aperture radar (SAR) data can offer insightful and systematic forest monitoring with complete time series (TS) due to signal penetration through clouds and day and night image acquisitions. This study aims to address the limitations of optical satellite data by using SAR data as an alternative for estimating optical VIs for forests through machine learning (ML). While this approach is less direct and likely only feasible through the power of ML, it raises the scientific question of whether enough relevant information is contained in the SAR signal to accurately estimate VIs. This work covers the estimation of TS of four VIs (LAI, FAPAR, EVI and NDVI) using multitemporal Sentinel-1 SAR and ancillary data. The study focused on both healthy and disturbed temperate forest areas in Czechia for the year 2021, while ground truth labels generated from Sentinel-2 multispectral data. This was enabled by creating a paired multi-modal TS dataset in Google Earth Engine (GEE), including temporally and spatially aligned Sentinel-1, Sentinel-2, DEM, weather and land cover datasets (MMTS-GEE). The inclusion of DEM-derived auxiliary features and additional meteorological information, further improved the results. In the comparison of ML models, the traditional ML algorithms, RFR and XGBoost slightly outperformed the AutoML approach, auto-sklearn, for all VIs, achieving high accuracies ($R^2$ between 70-86%) and low errors (0.055-0.29 of mean absolute error). XGBoost was found to be the most computationally effective and fastest algorithm. Great agreement was found for selected case studies in the TS analysis and in the spatial comparison between the S2-based and estimated SAR-based VIs. In general, up to 240 measurements per year and a spatial resolution of 20 m can be achieved using estimated SAR-based VIs with high accuracy. A great advantage of the SAR-based VI is the ability to detect abrupt forest changes with sub-weekly temporal accuracy.

## Introduction

Covering approximately 30% of the Earth's surface (Hansen et al. 2013), forest plays an inevitable role in the preservation of biodiversity, regulating global and regional carbon cycles (Pan et al. 2011; McMahon, Parker, and Miller 2010). Nowadays, in the era of big data and cloud-computing platforms, remote sensing (RS) data and techniques have become an essential part of forest monitoring from a local to global scale. They enable us to monitor the dynamics of large-scale forest ecosystems in a systematic way and almost in a near-real time. Monitoring forest changes using RS data and methods is crucial for tracking current greenhouse gas emissions, biodiversity indicators, or illegal logging activities and their consequences. Especially, on a regional scale it can help to provide a rapid adequate response in forest management in case of sudden changes. Recent studies have proven the key role of RS in monitoring changes in European temperate forests caused by various natural disturbances, such as bark beetle calamities (Hlásny et al. 2021; Lastovicka et al. 2020), windthrows (Dalponte et al. 2020; Baumann et al. 2014), wildfires (Quintano, Fernández-Manso, and Fernández-Manso 2018; García-Llamas et al. 2019) or drought (Schellenberg et al. 2023; Kaiser et al. 2022), but also forest regeneration and regrowth (De Luca, Silva, and Modica 2022b).

Forest behavior, especially in temperate forests, undergo seasonal changes that impact satellite observations. In the optical domain, these variations are attributed to alterations in the physiological, biophysical, and biochemical states of vegetation (De Luca, Silva, and Modica 2022b). Traditionally, optical vegetation indices (VIs) have been found to be effective in monitoring forest health conditions and its phenology phases. The most widely-used vegetation indices and biophysical parameters for vegetation monitoring are the Normalized Difference Vegetation Index (NDVI), Enhanced Vegetation Index (EVI), Leaf Area Index (LAI) and the Fraction of Absorbed Photosynthetically Active Radiation (FAPAR). These parameters are widely used in global products with various spatial resolutions spanning from 1000 to 10 m and on a temporal resolution of 16 days from VIIRS and MODIS missions (Didan 2021b; 2021a; Didan and Barreto 2018; Myneni, Knyazikhin, and Park 2021), 10 days from the Copernicus Global Land Service (Fuster et al. 2020) or on a daily basis through the Pan-European High-resolution vegetation phenology and productivity products derived from Sentinel-2 data (Smets et al. 2023). However, the real temporal resolution of these products depends on the availability of cloud-free acquisitions. In general, optical VIs are also prone to atmospheric effects, which can make even cloud-free measurements incomparable.

On the other hand, Synthetic Aperture Radar (SAR) data which are not prone to clouds and atmospheric effects and with daylight-independence can not only overcome the problems that affect optical data, but can also serve as a substitute in order to estimate the values of data gaps in optical RS. SAR data have been utilized to monitor deforestation and forest degradation in tropical areas (Reiche et al. 2021), bark beetle calamities (Marinelli et al. 2023), wildfire progression (Ban et al. 2020; Paluba et al. 2024; Di Martino et al. 2023), windthrow detection (Rüetschi, Small, and Waser 2019; Dalponte et al. 2023) and seasonal variation monitoring (Dostálová et al. 2016; 2018). In the SAR domain, forest changes are primarily driven by structural modifications in trees, such as shape, orientation, and leaf/needle density (De Luca, Silva, and Modica 2022a), in combination with environmental factors like temperature and precipitation (Benninga, van der Velde, and Su 2019; Rüetschi, Small, and Waser 2019; Olesk et al. 2015).

Based on the relationship found between optical and SAR polarizations and indices, some researchers tried to estimate optical VIs using SAR data and machine learning (ML) methods. In Filgueiras et al. (2019), 8 different regression algorithms were tested to estimate the NDVI for crop monitoring, from which the random forest regressor (RFR) performed the best with a mean absolute error (MAE) of 0.034-0.075 and a Root Mean Square Error (RMSE) of 0.020-0.037, outperforming more advanced methods like gradient boosting or the Bayesian-regularized neural networks. They also found that models incorporating local incidence angle (LIA), along with pure SAR features, achieved the highest accuracy. Next to NDVI, estimation of EVI with ML was tested in dos Santos et al. (2022), while finding the RFR as the best performing regressor. RFR as the best performing algorithm was found in Lasko (2022) and Mohite et al. (2020) for filling data gaps in NDVI images using SAR data and for NDVI prediction from SAR data for agricultural crops. Lasko (2022) found that seasonal variations impact the predictive capability of Sentinel-1 SAR, where the RFR was more effective during spring/summer seasons. They utilized a combination of C-band and L-band SAR data with GLCM texture and de-noising techniques, achieving high accuracy in predicting NDVI and NDWI, with $R^2$ values of 91% and 87%, respectively. In these works, tabular regression was used. When it comes to current advancements in tabular machine learning tasks, the gradient boosting algorithm, especially the Extreme Gradient Boosting (XGB), was reported to outperform other traditional methods, including RFR, while very well generating on different types of datasets (Grinsztajn, Oyallon, and Varoquaux 2022). XGB was recently found to outperform RF in classification tasks in urban applications (Shao, Ahmad, and Javed 2024) or to outperform even deep learning methods in regression tasks for crop yield prediction (Huber et al. 2022).

More advanced methods using image patches in a deep learning (DL) approach were used, e.g. in Roßberg and Schmitt (2022) and Roßberg and Schmitt (2023). They estimated NDVI for different land cover types globally using a U-net architecture with 2300 temporally and spatially paired S1 and S2 image patches, achieving a Mean Absolute Error (MAE) of 0.1. They also found that adding DEM or performing radiometric terrain correction does not improve the results using DL approaches, while on the other hand, inclusion of a land cover dataset significantly improved their results. In Calota, Faur, and Datcu (2022), the bag-of-words method in a three-layered and VGG-19 CNN was used to estimate the NDVI values from SAR data using BigEarthNet image patches. CNN to predict NDVI values was used in Mazza et al. (2018), who found that fusing temporally adjacent SAR images and adding a speckle filter improves accuracy.

Nowadays, when there is a rapid increase of new ML algorithms (Tuia et al. 2023; 2017; Audebert, Le Saux, and Lefèvre 2017), it can be difficult to decide which algorithms to include in the comparison and how to find the best performing ML pipeline for a specific task. Automatic Machine Learning (AutoML) approaches are out-of-the-box ML methods that aim at finding the best performing ML pipelines, selecting the best data and feature preprocessing methods, the best algorithms and their hyperparameters, and creating an ensemble of the best models. All of this is performed using an optimizer, such as a Bayesian optimizer, to boost the optimizations by learning from previous steps. Therefore, AutoML can automate the repetitive process of finding the best ML pipeline and also lowers the barrier to create high-performing models for less experienced users in ML. The RS community gained interest in various AutoML frameworks, such as auto-sklearn (Feurer et al. 2015), Tree-based Pipeline Optimization Tool (TPOT) (Olson and Moore 2019) or AutoGluon (Erickson et al. 2020) for tabular-based tasks or Auto-Keras (Jin et al. 2023) for image-based tasks, e.g. DL classification. Current RS-oriented works using AutoML frameworks present various applications, e.g. for preparation of an AutoML classification design for RS data (Palacios Salinas et al. 2021), for classification of a tropical landscape (Kiala et al. 2020), for identification of weeds (Espejo-Garcia et al. 2021), landslide susceptibility (Renza et al. 2021). For regression tasks, AutoML was used for tomato yield prediction (Darra et al. 2023), modeling forest aboveground biomass (Naik, Dalponte, and Bruzzone 2023), estimating grass height (César de Sá et al. 2022) or predicting grape sugar content (Kasimati et al. 2022). Auto-sklearn is one of the most used AutoML frameworks in general, but also in RS. It is open-source and is completely built on the popular scikit-learn Python library (Pedregosa et al. 2011). Furthermore, in benchmark tests, auto-sklearn outperformed other popular AutoML frameworks for both classification in Feurer et al. (2015) and Feurer et al. (2022), and regression tasks in Conrad et al. (2022), while better performance compared to AutoGluon was also achieved in a regression using RS data in César de Sá et al. (2022).

The goal of this study is to estimate optical VI time series (TS) for forest monitoring using SAR and other ancillary data (e.g., elevation, weather conditions) with ML approaches. For this purpose, in the first step, a temporally and spatially paired dataset of S1, S2 and ancillary data is generated. Two traditional and state-of-the-art tabular ML approaches (RFR and XGB) are then compared to the auto-sklearn as an AutoML approach. Moreover, the performance of auto-sklearn was tested with different optimization lengths. The effect of additional topographic and weather features are evaluated as well. The advantages and drawbacks of the methods used and the SAR and other ancillary data are discussed in the estimation of the TS of optical VIs.

# Materials and Methods
## Data

Only open-access data were used in this study, with the main focus on Sentinel-1 SAR data (S1) and Sentinel-2 Level-2A multispectral data (S2), both assessed in the GEE platform. The dual-polarized Copernicus S1 SAR Ground-range detected (GRD) data operate in the C-band with a 5.405 GHz central frequency. The native spatial resolution of these data are about 20 x 22 m; therefore, a spatial resolution of 20 m was used in data processing and analyses. Due to an anomaly that occurred to the Sentinel-1B satellite at the end of 2021 (Pinheiro et al. 2022), S1 data acquired before 2022 were used to explore the full potential of both S1 satellites providing a 6-day temporal resolution. The S2 mission has currently two sun-synchronously orbiting satellites providing images in the optical domain with a spatial resolution of 5 days with a spatial resolution of 10, 20 and 60 m, depending on the spectral band.

To test the usefulness of ancillary data, the Copernicus DEM digital elevation model, specifically the global C-DEM GLO-30 product (C-DEM), and weather data from the ERA-5 Land mission were included in the regression analysis. Elevation models provide useful information on altitude which may determine forest composition and density, and have a proven record of improving ML algorithms (Carvalho et al. 2020; Hänsch et al. 2022). The ERA-5 Land dataset demonstrated the second-best accuracy over other precipitation products for the Central European environment, Czechia, with the top-performing dataset providing both precipitation and temperature data with time series dating back to 1950s (Paluba, Blizňák, et al. 2024). The C-DEM digital surface model with a spatial resolution of 30 m is based on TanDEM-X SAR images acquired between 2011 and 2015 (European Space Agency and Airbus 2022). The C-DEM achieved superior accuracy compared to other open-access global DEMs in current comparison (Bielski et al. 2024; del Rosario González-Moradas et al. 2023; Li et al. 2022; Marešová et al. 2021).

The relationship of temperature and precipitation with C-band SAR variables was found in previous studies, such as in Benninga, van der Velde, and Su (2019); Olesk et al. (2015); therefore, two weather variables of the ERA-5 Land were used: 'temperature_2m' – an average temperature of air 2 meters above the surface; and 'total_precipitation' – accumulated liquid and frozen water, including rain and snow, that falls to the Earth's surface. The ERA5-land reanalysis dataset has an hourly temporal and a 0.1 arc degrees spatial resolution (equivalent to a native resolution of 9 km) (Copernicus Climate Change Service 2019, 5).

# Study Areas

## Study Areas: Input Dataset Generation and Verification for Training, Validation and Testing

Forest areas in Czechia were selected as study areas. The input dataset included 600 healthy coniferous, 600 healthy broad-leaved and 600 disturbed coniferous areas with a 20x20 m area. Healthy forest areas generated and validated in Paluba, Le Saux, et al. (2024) were used. Disturbed forest areas were also added to introduce greater variation in the input dataset. They were generated by, first, selecting pixels indicating forest loss between 2018 and 2021, as identified in the GFC dataset. Then, around 1200 with a 20 m buffer were selected to represent areas entirely covering the disturbed coniferous forest masks. These areas were further validated using high resolution imagery in Google Earth Pro and a final set of 600 disturbed coniferous areas were preserved for further analysis. The differentiation between coniferous and broad-leaved forests was based on an intersection of the loss mask, CLC from 2018 and GLCL from 2019. Disturbed broad-leaved forest areas were excluded from the analysis, as only a few examples were found according to the above-mentioned requirement. The spatial distribution and terrain characteristics of the input dataset can be found in Fig. 1 and Table I, respectively.

Table I. Terrain-based characteristics of the input dataset

| Input class type | Mean/median altitude in meters (Q1-Q3) | Mean/median slope in degrees (Q1-Q3) | Mean/median LIA in degrees (Q1/Q3) |
|---|---|---|---|
| Coniferous forests | 650/643 (493-793) | 10/8 (5-13) | 39/39 (33-45) |
| Broad-leaved forests | 417/406 (304-505) | 12/11 (6-16) | 38/39 (32-45) |
| Disturbed coniferous | 580/587 (509-649) | 9/8 (5-12) | 38/38 (32-43) |

To prepare training and testing datasets, 30% of the samples from each class in the input dataset (healthy broad-leaved, healthy coniferous and disturbed coniferous forests) were randomly selected as testing data, and the remaining 70% were used in training. This division was done separately for each class, also ensuring that samples from the same area (with the same *ID*) could not be included in both the training and testing dataset. A new categorical feature, called "forest_type", was added, which specifically distinguishes between broad-leaved and coniferous forests. This is important because broad-leaved and coniferous forests have different phenological responses/seasonality over the year and it is also detectable using TS of both SAR and optical indices, e.g., in Dostálová et al. (2016).

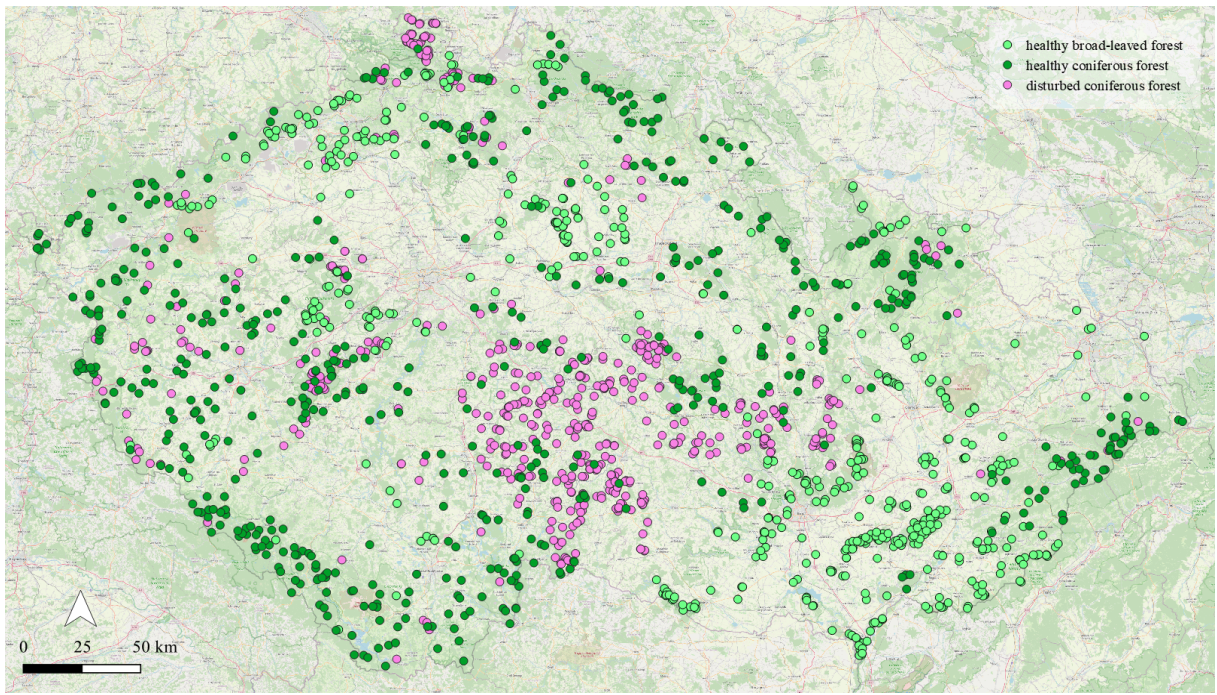

Fig. 1. Distribution of the forest areas used in this study.

Case Studies for Healthy and Disturbed Forest TS Analysis

For qualitative and quantitative comparison of the optical VIs based on S2 TS and the estimated SAR-based VIs (detailed in the subsection named "Accuracy Assessment and Visual Analysis of the Results"), four case studies with a 40x40 m area were selected. For healthy case studies, one broad-leaved and one coniferous forest area was selected. To demonstrate the ability of the model to recognize and predict disturbed areas, one area where the disturbance occurred during the monitored time frame, i.e., in 2021, and one area where the disturbance occurred before 2021, were selected. General information, such as the X, Y coordinates of the center, tree cover composition, terrain characteristics and the number of available SAR acquisitions in 2021 can be found in Table II.

Table II. Characteristics of the selected case studies.

| Forest type | X, Y coordinates of the center | Tree cover | elevation (m) / slope (°) / LIA range (°) | # SAR images |
|---|---|---|---|---|
| Broad-leaved | 49.95776, 14.17296 | 100% tree cover | 432 / 3 / 34-47 | 236 |
| Coniferous | 50.44805, 14.45740 | 100% tree cover | 271 / 4 / 32-44 | 176 |
| Disturbed coniferous during 2021 | 49.82402, 15.11192 | 100% tree cover to clear-cuts | 552 / 9 / 23-48 | 232 |
| Disturbed coniferous before 2021 | 49.35071, 15.51560 | 100% clear-cuts | 641 / 4 / 30-39 | 175 |

Test area for spatial comparison of results

For visual comparison of the S2-based and SAR-based VIs, a test area in the north-west of Czechia was selected, in the Krušné hory mountains. For this purpose, the NDVI was selected. The selected area represents a complex mountainous terrain with a mean elevation of 742 m above sea level and mean slopes of 13°. It has almost the same representation of coniferous and broad-leaved forests (45% to 55% share) and contains around 28,000 forest pixels of 20 m spatial resolution (around 11 km$^2$). The S2 optical and S1 SAR acquisitions are from the same day (8th September) with a temporal difference of 7 hours. This area was selected to represent mostly forest areas, where forest types differentiation and non-forest area masking was performed using the Copernicus High Resolution Layers (HRL) (Langanke 2017). Characteristics of the test area can be found in Table III.

Table III. Characteristics of the test area. C stands for coniferous forests; B stands for broad-leaved forests.

| Name | Forest pixels (in thousands) | Forest type (C / B) in % | mean elevation (m) / slope (°) / LIA (°) | Date and time of S2 and S1 acquisitions |
|---|---|---|---|---|
| Krušné hory test area | 28 | 45 / 55 | 742 / 13.0 / 35.1 | S2: 08/09/2021 10AM S1: 08/09/2021 5PM |

## Data Pre-Processing and Feature Creation

In the first step, data were assessed and pre-processed in the GEE platform. Sentinel-1 and Sentinel-2 image collections were filtered to select each acquisition in 2021 for the study area, Czechia. S1 GRD data in GEE are already preprocessed using an orbit file application, GRD border noise removal, thermal noise removal, radiometric calibration and terrain correction (Google Earth Engine 2024). It should be noted that radiometric terrain correction was not applied. Moreover, a Lee speckle filter with window size of 5x5 (Lee 1985) was used to eliminate the effects of speckle noise using an implementation of Mullissa et al. (2021). VV/VH and VH/VV polarimetric indices were calculated to enhance the SAR feature space. These features were selected along the VV and VH polarizations, based on the findings in Paluba, Le Saux, et al. (2024), where these features were found to be the most suitable for forest monitoring in the Central European environment. S2 data were cloud masked using the CloudScore+ approach (Pasquarella et al. 2023) using the 'cs' band and using a 0.6 threshold. Cloud masking was followed by vegetation index derivation, especially the NDVI, EVI, LAI and FAPAR were derived. LAI and FAPAR were derived using the GEE implementation developed by Van Tricht (2023).

To enhance the feature space, elevation, slope and local incidence angle (LIA) were derived from the C-DEM. The LIA was calculated for each S1 image separately based on Paluba et al. (2021). The ERA-5 Land 'total_precipitation' variable was used to derive the sum of precipitation 12 hours prior to each SAR acquisition, while the 'temperature_2m' variable was used to obtain the air temperature at the time of each SAR acquisition. A paired multi-modal TS of S1 and S2 data, with a maximum 24 hour temporal difference between S1-S2 image pairs, extended with DEM and weather features, was created using the multi-modal TS generation tool in GEE (MMTS-GEE), based on Paluba, Le Saux, et al. (2024).

Preliminary data cleaning and preparation was performed prior to any analysis. Records which included null / missing values were excluded from the analysis, thus only records having values in each feature were included in the further analysis. The acquisition dates in day of the year format (DOY) were transformed into cyclical sine and cosine features to ensure the proximity of values obtained in January to those obtained in December.

The following set of input features were considered in the main analysis: VV, VH, incidence angle (angle), VV/VH, VH/VV as SAR features; LIA, elevation and slope as DEM features; sum of precipitation 12 hours prior to SAR acquisition ($prec._{12h}$) and temperature at the time of SAR acquisition (temp.) as weather features; the forest type as a differentiating feature between coniferous and broad-leaved forests; and $DOY_{sin}$ and $DOY_{cos}$ containing information about the time of the corresponding SAR acquisition.

To test the effect of ancillary features on the results, the effect of input features was tested for each VI separately. The following input feature sets were tested using the best performing models: 1) solely SAR features, 2) SAR and DEM-based features and 3) using all above mentioned features, that is SAR, DEM- and weather-based features.

## Forest Parameter Estimation with traditional ML and AutoML

Regression analysis was used in this study. The main part of the analysis consisted of comparing the performance of an AutoML approach, auto-sklearn, and traditional ML approaches, RFR and XGB, in estimating optical VIs from SAR and ancillary data for forests.

### Random Forest Regression and Extreme Gradient Boosting, as traditional ML approaches

The RFR and the XGB were selected as baseline algorithms in this study due to their superior performance in VI estimations, optical data gap filling using SAR data or in recent tabular ML tasks. The RFR is one of the most used ML algorithms for classification and regression tasks in RS and it operates as an ensemble of decision trees, where each tree is constructed using randomly selected subsets of training datasets and features (Belgiu and Drăguţ 2016). Gradient boosting algorithms are ensemble learning methods that utilize iterative learning, where errors from previous iterations are considered in the improvements of the subsequent iterations. In addition, the XGB algorithm is an optimized implementation of regular gradient boosting algorithms by regularization to reduce overfitting (e.g. through early stopping), parallel processing for faster computation or missing data handling, making it more efficient and robust for large-scale data (Chen and Guestrin 2016). Additionally, both RFR and XGB have shown outstanding performance in classification tasks across 140 datasets compared to other classifiers and auto-sklearn, as reported in Feurer et al. (2015). The Scikit-Learn Estimator Interface of the *xgboost* Python package (github.com/dmlc/xgboost, last accessed on 28.7.2024) and the RFR model from the scikit-learn library were used in this work.

Both models were first fine-tuned to find the best-performing hyperparameter combinations using each 14 input features and on each VI separately. The RFR fine-tuning was focused on number of trees (*n_estimators*: from 50 to 500 with a step of 50) and the number of features to consider when looking for the best split (*max_feautres*: 'sqrt', 'log2' and each value from 1 to 14) with a loss function set to MAE (*criterion* = 'mean_absolute_error'). The fine-tuning of the XGB was focused on the maximum tree depth (*max_depth*) and learning rate (*learning_rate*) parameters, while the number of gradient boosted trees (n_estimators) was set to 5000, while the early-stopping usually stopped the tree growth much earlier (between 20 and 1000 depending on the learning rate). The evaluation metric for both early-stopping (*eval_metric*) and as a general loss function (*objective*), was set to MAE and AE, respectively. The use of categorical features was enabled due to the *forest_type* feature. To reduce the overfitting, early-stopping was activated, where the validation metric needs to improve at least once in every 5 rounds to continue the training, otherwise it stopped.

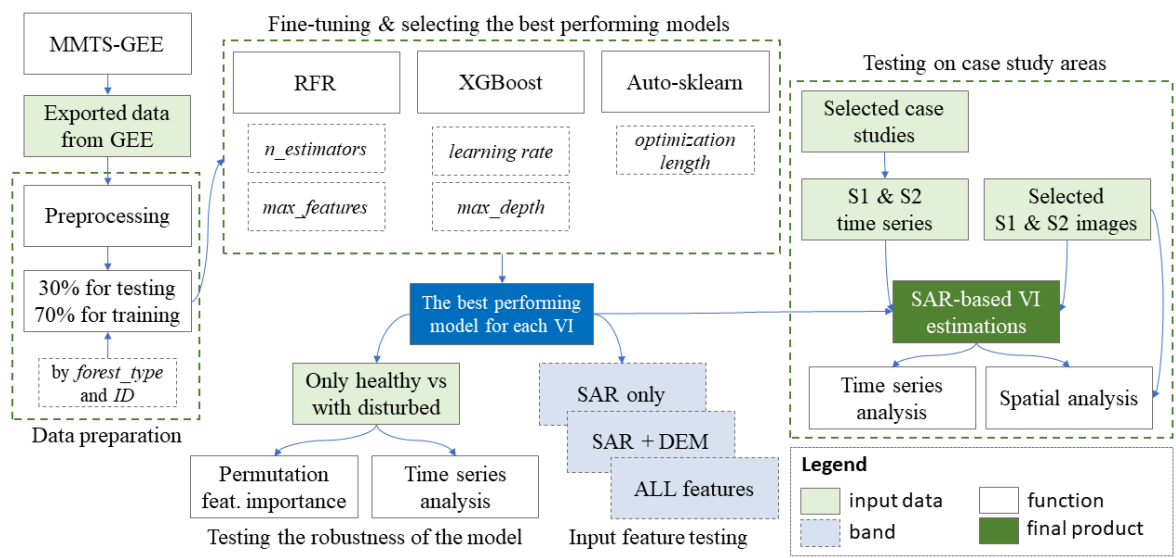

Fig. 3. Data analysis workflow.

### Auto-sklearn, as an AutoML approach

An AutoML tool can be described as an out-of-the-box supervised ML approach that enables automatic development of ML pipelines. Auto-sklearn is able to perform all the data and feature pre-preprocessing, as well as the algorithm selection and its hyperparameter tuning for the user. The process of identifying the best performing algorithms and their hyperparameters, known as Combined Algorithm Selection and Hyperparameter optimization (CASH), is enabled by Bayesian optimization, specifically the Sequential

Model-Based Optimization for General Algorithm Configuration (SMAC) (Hutter, Hoos, and Leyton-Brown 2011), which is based on random forest models (Thornton et al. 2013). On top of solving the CASH problem, auto-sklearn also uses meta-learning to warm-start the Bayesian optimization process by identifying similarities in the pre-trained datasets (details in Feurer et al. 2015), and thus it significantly speeds up the optimization process. Bayesian optimization ensures that the algorithm can learn from the previous steps in the loop and suggest a better setting (pre-processors, algorithms, hyperparameters) in the next step. At the end, it builds an ensemble of ML models considered by Bayesian optimization. The auto-sklearn ML pipeline optimization is illustrated in Fig. 4. For regression tasks, auto-sklearn includes 12 regression algorithms, 15 feature preprocessing algorithms and 5 data preprocessing algorithms, all based on the scikit-learn library (Pedregosa et al. 2011, 525). Altogether, there are 95 hyperparameters which can be evaluated in pipeline optimization, while most of them are conditional hyperparameters (active when another hyperparameter is active) (Feurer et al. 2015). Each algorithm with the number of their hyperparameters is listed in Appendix I. Auto-sklearn also supports multi-output regression analysis, but it is allowed only for five available regression models (K-nearest neighbor, Decision tree, Extra Trees, Gaussian Process, Random Forest). Therefore, to include as many regressors in the search space as possible, the analysis was performed for each target optical VI separately.

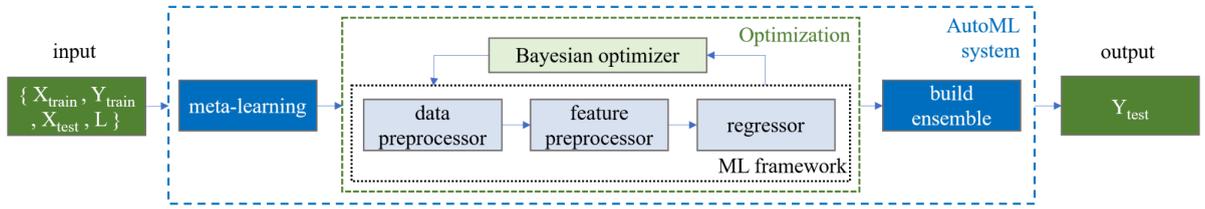

Fig. 4. Auto-sklearn search space pipeline for regression analysis. Based on (Feurer et al. 2015).

The pipeline optimization process is performed until a defined computational budget, which in case of auto-sklearn is time. A longer time usually means a higher chance of finding better-performing pipelines. To test the required optimization length for auto-sklearn, in this study, four different optimization lengths (10 minutes, 1, 6 and 12 hours) were tested. The user can also define how much time can be left for a single model to be performed. In this study, the default value representing one tenth of the overall time limit (the overall computation budget) left for AutoML was used. As a metric to evaluate the final ensemble model (loss function - L), the MAE was used to eliminate the absolute error in the prediction.

It should be noted that the processing power settings used for training, fine-tuning of RFR and XGB and for optimization of auto-sklearn were the following: 12th Gen Intel(R) Core(TM) i7-12700 with 2.10 GHz, 64 GB of RAM and 20 CPU cores.

Accuracy Assessment and Visual Analysis of the Results

Successful and unsuccessful runs were examined after each run of auto-sklearn. The ensembles of the best found models, the weight of each individual component in the ensemble, the duration and cost were saved.

The best found ensemble model was used in inference for the test datasets. The accuracy was evaluated using three metrics: MAE (1), Root Mean Squared Error (RMSE) (2) and for the comparison of accuracies between different models, the Coefficient of determination, $R^2$ (3) was used. They are defined as follows:

$$\text{MAE} = \frac{1}{n} \sum_{i=1}^{n} |y_i - p_i| \qquad (1)$$

$$\text{MSE} = \sum_{i=1}^{n} (y_i - p_i)^2 \qquad (2)$$

$$R^2 = \frac{\sum_{i=1}^{n} (y_i - p_i)^2}{\sum_{i=1}^{n} (y_i - \bar{y})^2} \qquad (3)$$

where $y_i$ stands for the i-th actual (label) value, $p_i$ stands for the predicted/estimated value, $\bar{y}$ is the calculated average of the actual values $n$ is the sample size (the size of the test set).

Because of the stochastic nature of ML algorithms, 3 different runs of auto-sklearn were performed for accuracy assessment. The average of accuracy scores of these runs were calculated for each estimated VI and for each length of auto-sklearn optimization runs. The stochastic nature of ML refers to the introduction of randomness into the learning algorithms. This stochasticity is often used to improve the performance and generalization of ML models.

## Testing the Robustness of the Model, Time Series and Spatial Analysis

To underscore the importance of data variability when estimating optical VIs and to test the robustness of the model, two input datasets were tested: 1) data representing solely healthy forest areas and 2) data including disturbed forest areas. In this experiment, all input features were used for estimating NDVI with the best performing model. The permutation feature importance and the behavior of the TS on case studies was assessed for both input datasets on the testing dataset. The permutation feature importance measures the contribution of each feature to the overall performance of the model. This analysis aimed to explore how input features other than SAR features might affect the regression results, considering that excluding disturbed forest areas could reduce the variability in SAR features.

Once the best regressors were chosen for each VI, a TS analysis was performed in selected forest areas (case studies) to compare the S2-based and predicted SAR-based VI TS for healthy broad-leaved, coniferous and disturbed coniferous forests. To enhance the visualizations, an averaging moving window of size 5 was applied to the TS data. This helps smooth out the data and provides clearer representation of the trends and patterns in VI values over time.

Furthermore, a spatial comparison of the S2-based and estimated SAR-based NDVI was performed for the test area in Krušné hory mountains. A map of absolute errors was also produced, while the MAE and the standard deviation for the selected test area was calculated. As the forest type is included in the input dataset for training, forest type differentiation (coniferous versus broad-leaved forests) is needed for inference as well. The CLC and CGLCL layers have a 100 m minimal mapping unit; therefore, a more detailed dataset was used for this inference, the Copernicus High Resolution Layers (HRL) from 2018, which is generated in 10 m spatial resolution.

# Results
## Fine-tuning of the RFR and XGB

To find the best hyperparameter combination for RFR and XGB, fine-tuning was performed using all input features for each VI separately. Generally, for RFR an increase of the *max_features* hyperparameter until 7-10 improved the results, while 1 revealed the worst results. Similarly, the increase in the number of trees (*n_estimators*) caused a decrease in MAE. Slight variations in the best found hyperparameters were observed between the top 20 combinations and the overall best combination, with *max_features* ranging between 7-8 and 4 for EVI and *n_estimators* fluctuating between 150-500 (see Fig. 5a for NDVI and the Appendix II for details on each VI).

Table IV. Best hyperparameter combinations identified for RFR and XGB. Bolded results represent the best achieved results for the VI.

|  | RFR | | | | XBG | | | |
| --- | --- | --- | --- | --- | --- | --- | --- | --- |
| VI | n_estimators | max_features | MAE | time (s) | learning rate | max_depth | MAE | time (s) |
| **LAI** | 500 | 8 | **.2938** | 227.91 | 0.01 | 9 | .2970 | **3.16** |
| **FAPAR** | 500 | 7 | **.0736** | 208.25 | 0.01 | 11 | .0742 | **5.26** |
| **EVI** | 350 | 4 | .0487 | 89.07 | 0.02 | 9 | **.0480** | **1.58** |
| **NDVI** | 450 | 8 | .0751 | 226.79 | 0.01 | 12 | **.0734** | **8.07** |

The XGB fine-tuning revealed a general decrease in MAE for lower learning rates (0.10-0.01) and *max_depth* between 8 and 12 for each VI (see Fig. 5b for NDVI and the Appendix II for details on each VI). Higher learning rates ended up with higher MAE, while the worst results were obtained with *max_depth* < 4. Similarly as for RFR, slightly different hyperparameters were found as best for each VI (Table IV), therefore, these best hyperparameter combinations were used in the next analyses. The best found hyperparameter combinations for RFR and XGB for each VI are detailed in Table IV. The fitting time for XGB was about 30- to 70-times faster (2-8 seconds) compared to the fitting time of RFR (89-200 seconds).

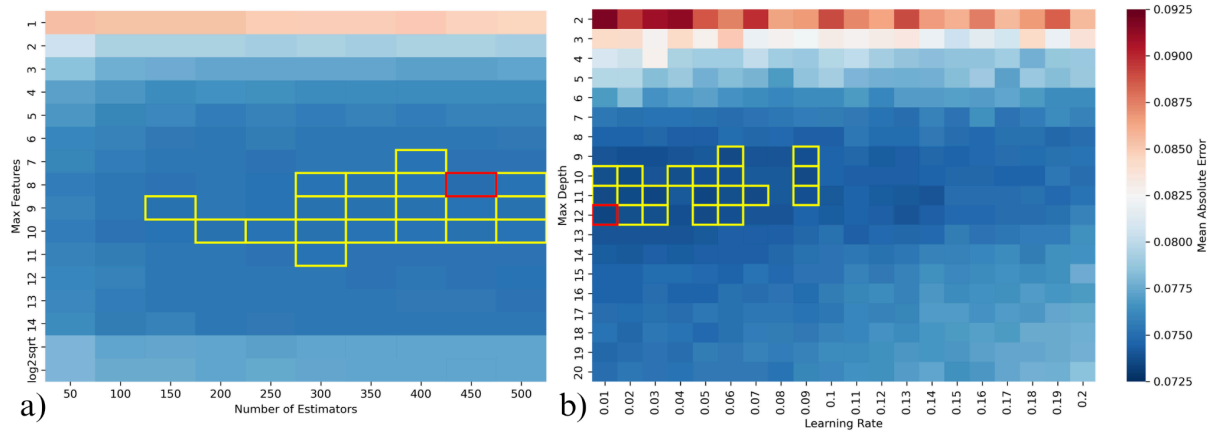

Fig. 5 Fine-tuning results for RFR (a) and XGB (b). The top 20 results with the lowest MAE are highlighted with yellow, the red highlighted cell represents the best performing hyperparameter combination.

# Evaluation of Auto-sklearn

In this step, the performance of Auto-sklearn was assessed for different training lengths. A significant improvement can be found mainly when increasing the optimization length from 1 to 5 minutes. The highest improvement in $R^2$ was made to FAPAR (by 63%), while for NDVI and LAI it was by 15% and 5%, respectively. On the other hand, EVI experienced a decrease in $R^2$ and increase of the MAE and MSE when increasing the optimization length from 1 to 5 minutes. In case of EVI, the highest accuracy was achieved with a 1 minute optimization time, where 3 GB algorithms were found. It was found that the first peak performances were achieved when only GB formed the final ensembles: at 1 minute for EVI, at 10 minutes for FAPAR, 15 minutes for LAI and 60 minutes for LAI. In case of LAI and EVI, these results represented the best $R^2$ achieved for Auto-sklearn optimizations, that is 85.57% for LAI at 15 minutes with or EVI at 1 minute with 81.63%. The highest $R^2$ for FAPAR and NDVI were achieved at 6 hours with 77.39% (only 0.03% increase compared to 10 minutes length), and at 12 hours with 71.47% (only 0.86% increase compared to 60 minutes length), respectively. Therefore, generally, the accuracy was not increasing or increasing just slightly with increasing the AutoML optimization length.

With the increasing time budget for auto-sklearn optimization length, an increase in the number of tested pipelines and the share of successfully finished pipelines can be detected (Fig. 6). This is an expected behavior of the AutoML approaches. In the case of 1-minute runs, on average, 39 pipelines were tested, while only 20% of these pipelines were finished successfully. In the 10-minute runs, on average, 92 pipelines were tested and 40% were successful, while in 1-hour runs the success rate was 56% out of 129 tested runs on average. In 6 and 12 hour runs, 70% and 73% of the pipelines ended successfully from the total of 229 and 324 evaluated pipelines, respectively.

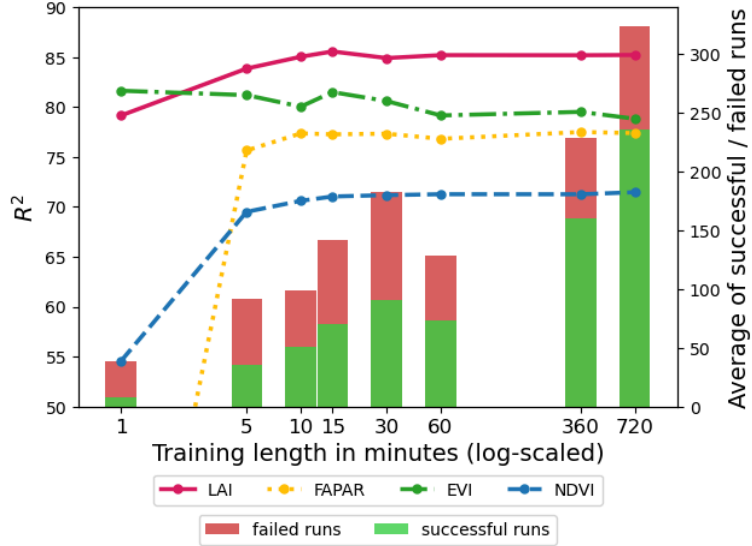

Fig. 6. Evaluation of results using input data using all available features with auto-sklearn regression for different lengths of optimization runs, each averaged across 3 repetitions.

## Selecting the best performing regressor

At the end, the best performing regressors based on MAE were identified for each VI. It should be noted that the best AutoML results based on MAE were achieved with an 1 hour optimization length for LAI, 6 hours for FAPAR and EVI and 12 hours for NDVI. In all cases, RFR or XGB slightly outperformed the AutoML pipeline almost for each VI (Table V), while AutoML has not achieved the best results for either VI. The best results are therefore formed by XGB or RFR models. For LAI and FAPAR, the RFR with the 500 decision trees and with 8 and 7 *max_trees*, respectively, were selected. For EVI and FAPAR, the XBG with learning rates of 0.02 and 0.01, and max_depth of 9 and 12 were found to be the best performing, respectively. The $R^2$ ranged between 70.30% and 86.75%, while LAI achieved the highest $R^2$.

Table V. Best hyperparameter combinations identified for RFR and XGB, and best found results for AutoML. Bolded results represent the best achieved results for the VI.

|       | RFR      |       |       | XBG     |       |       | AutoML  |       |       |
|-------|----------|-------|-------|---------|-------|-------|---------|-------|-------|
| VI    | $R^2$ (%) | MSE   | MAE   | $R^2$ (%) | MSE   | MAE   | $R^2$ (%) | MSE   | MAE   |
| LAI   | **85.75** | **.1580** | **.2940** | 85.30 | .1630 | .2970 | 85.58 | .1598 | .2962 |
| FAPAR | **78.13** | **.0101** | **.0736** | 77.20 | .0105 | .0742 | 77.68 | .0103 | .0747 |
| EVI   | 78.66 | .0073 | .0496 | **83.38** | **.0057** | **.0480** | 80.16 | .0068 | .0495 |
| NDVI  | 70.95 | .0120 | .0751 | 70.30 | .0123 | **.0734** | 71.75 | **.0117** | .0735 |

## Proving the Robustness with the Inclusion of Disturbed Forest Areas

A comparison of model performances (both statistically and visually in TS analysis) and feature importances in NDVI estimation of the XGB model (as the best performing model) trained with only healthy forests as well as when including disturbed forests is presented here. When using only healthy forest areas as input dataset (for training, validation and testing), the XGB was able to predict NDVI with an $R^2$ of 85.11%, MAE of 0.043 and with a very good match of estimated NDVI TS and S2-based NDVI from S2 (Fig. 7a and b). However, in this case, the time features ($DOY_{sin}$ and $DOY_{cos}$) represented more than 62% importance in the permutation analysis, and the forest-type had about 21%. Other features had only a marginal effect on the results (<5%), while all SAR features had less than 1.3% (Fig. 8a). This means it would be enough to have only time and forest type information to estimate TS of the NDVI for healthy forests. On the other hand, the estimation of NDVI for the case studies of disturbed forests (Fig. 7c and d) shows almost identical behavior to estimation of healthy coniferous case studies. It is due to the fact that these disturbed areas represented coniferous forests and had assigned coniferous type in the inference.

When a higher variance was included in the input dataset, that is, when TS of 600 disturbed forest areas were added, the $R^2$ decreased to 70.30% and the MAE increased to 0.073. Although the statistical results became worse, the visual comparison of S2-based and estimated SAR-based VI TS still shows a very good match for healthy forest areas (Fig. 7a and b). More importantly, the SAR-based VI was able to detect the change (disturbance event) through its decrease at the DOY range of 201-204, while showing a good match with the S2-based NDVI (Fig. 7c). Lower values compared to healthy case study and a similar TS behavior to the S2-based observation were observed for the case study area, which was previously disturbed (Fig. 7d). The permutation importance (Fig. 8b) shows that the forest_type feature is the most important feature with 23.63%, while time features are still among the most important features with 39.6% contribution to the overall performance. The VH polarization had a contribution of 12.2%, while elevation and temperature had 6.5% and 5.5%, respectively. The other features had a lower than 5% contribution to the performance of the model, with VV polarization and both polarimetric ratios VV/VH and VH/VV achieving the lowest importance of 0.4-1.9%.

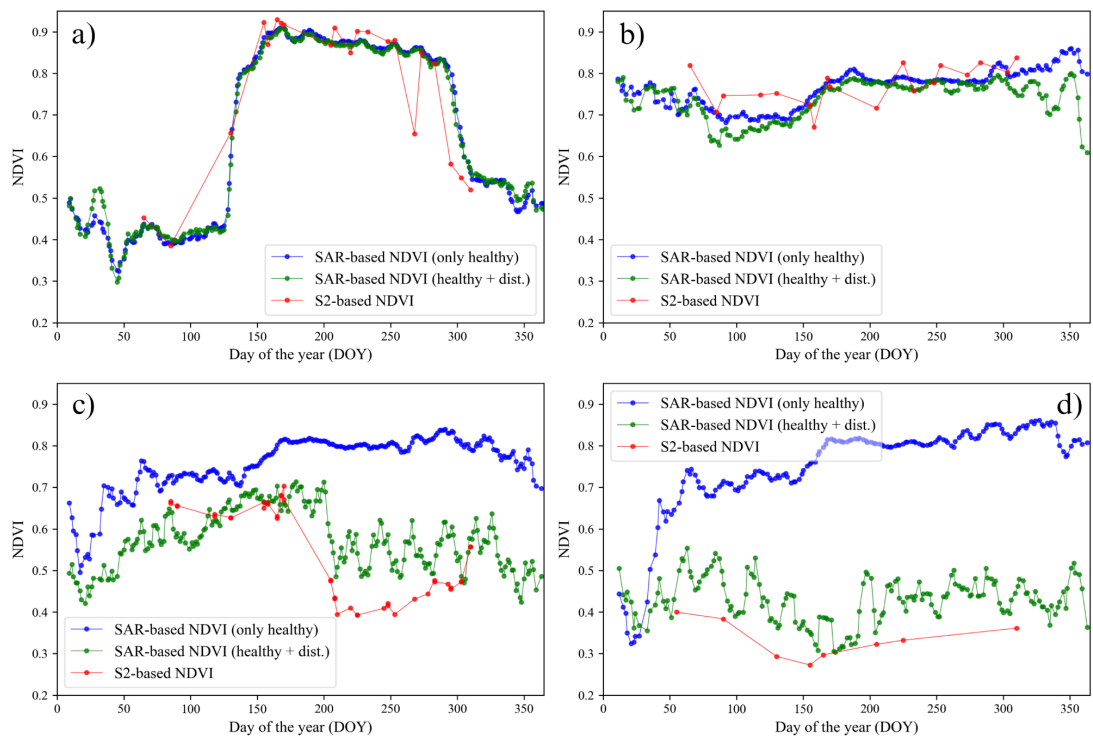

Fig. 7. TS for S2-based NDVI and predicted SAR-based NDVI when using only healthy forests in the training and predictions when using both healthy and disturbed forests in the training for healthy deciduous (a), healthy coniferous (b), disturbed coniferous forest areas (c and d).

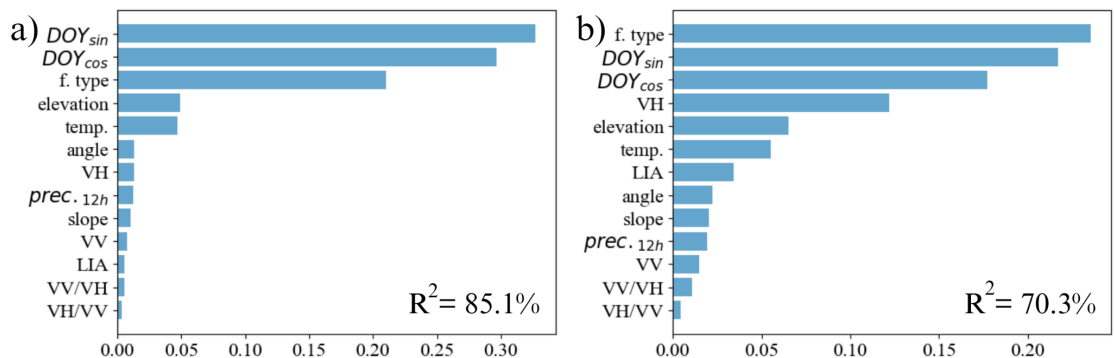

Fig. 8. Permutation importance when using only healthy forests (a) or using healthy and disturbed forests (b).

# Testing the Effect of Different Ancillary Features

The comparison of different input feature sets containing only SAR features, SAR with DEM-based features and using all input features (SAR with DEM- and weather-based features) using the best performing models is presented in Table VI. This comparison showed that using all input features reveals the best results in all accuracy metrics and for each VI. It should be noted that forest type and time information ($DOY_{sin}$ and $DOY_{cos}$) were included in each feature combination. The inclusion of features derived from the DEM (elevation, slope and LIA), represented a 5.6%, 4.6%, 2.7% and 1.2% increase in $R^2$ for NDVI, FAPAR, LAI and EVI, respectively, compared to estimations with just pure SAR data (details are in Table VI). Using additional weather information (precipitation 12 hours prior to the acquisition and temperature at the time of the acquisition), an additional improvement was made, specifically between 0.2% and 1.4% in $R^2$, while the greatest improvement was achieved for NDVI.

Table VI. Evaluation of results using input data with only SAR features, SAR and DEM, and using all available features for the best performing models. Bolded values represent the best achieved results.

| Feature type | only SAR | | | SAR + DEM | | | All features | | |
|---|---|---|---|---|---|---|---|---|---|
| | MAE | MSE | $R^2$ (%) | MAE | MSE | $R^2$ (%) | MAE | MSE | $R^2$ (%) |
| **LAI** | .330 | .195 | 82.37 | .301 | .166 | 85.07 | **.294** | **.158** | **85.74** |
| **FAPAR** | .083 | .013 | 72.39 | .076 | .011 | 76.97 | **.074** | **.010** | **78.21** |
| **EVI** | .050 | .006 | 81.99 | .049 | .006 | 83.22 | **.048** | **.006** | **83.38** |
| **NDVI** | .082 | .015 | 63.31 | .075 | .013 | 68.86 | **.073** | **.012** | **70.30** |

# Time Series and Spatial Analysis of the Results

TS analysis was performed for selected case studies from Table II, comparing S2-based VI TS and SAR-based TS. Furthermore, inference for the entire test area from Table III was performed to analyze the spatial compactness of the results. The TS analysis for selected case studies shows generally a good agreement between the S2- and the SAR-based TS for each selected forest type (broad-leaved, coniferous, disturbed coniferous during 2021 and disturbed coniferous forest before 2021) and shows a generally good differentiation between forest types with the ability to estimate the time of the disturbance event (Fig. 9). For the selected case studies, 9-21 measurements were available from the S2 mission, while from the S1 SAR mission, it was 175-236 depending on the overlap in acquisition paths over the selected area. The SAR data TS covered the whole year 2021.

Similarly to the S2 TS of the selected indices, the differentiation between broad-leaved (Fig. 9 green curves) and coniferous forests (Fig. 9 blue curves) is straightforward using the estimated SAR-based VI. This is due to the low values of the indices during the leaf-off period and a high value for the leaf-on period for broad-leaved forests and the generally stable behavior of coniferous forests throughout the year. The lowest VI values were achieved for the former coniferous forest area where a disturbance event occurred before 2021 (Fig. 9 red curves). An overestimation of SAR-based VIs can be detected for this case study, especially during winter. The clear drop in the S2-based VI TS for case study 3 is detectable between DOYs 170 (June 19) and 204 (July 23) (Fig. 9 dashed orange curve). On the other hand, in the SAR-based VI estimation (Fig. 9 orange curve), the drop is detectable between DOYs 201 and 204 (20-23 July 2021) for each SAR-based VI, with a slight increase of fluctuation in TS of VIs after the disturbance event. The fluctuation of the estimated TS for both disturbed areas is generally high, even if the average moving window of size 5 was used in Fig. 9. Furthermore, compared to the healthy coniferous case study, clearly lower VI values can be found in the pre-disturbance state in case study 3, suggesting that the condition of the forest was not in perfect shape before the disturbance.

In the spatial analysis of the estimated SAR-based NDVI, one can easily differentiate between coniferous and broad-leaved forest areas due to their difference in magnitude of the NDVI (higher NDVI for broad-leaved forests). Disturbed or non-forest areas are represented by low NDVI values both in the S2-based and SAR-based NDVI estimation. The spatial analysis also shows great agreement between SAR-based VIs and the S2-based

VIs, with a MAE of 0.061 and a standard deviation of 0.053 (Fig. 10). The highest errors were found in areas where recent forest disturbance occurred (mainly fresh clear-cuts in 2021), not captured by the HRL dataset from 2018, particularly in the central and south-east part of the study area. Higher errors can be found on the borders of between forested and non-forested areas, as well as on the borders of coniferous and deciduous forests.

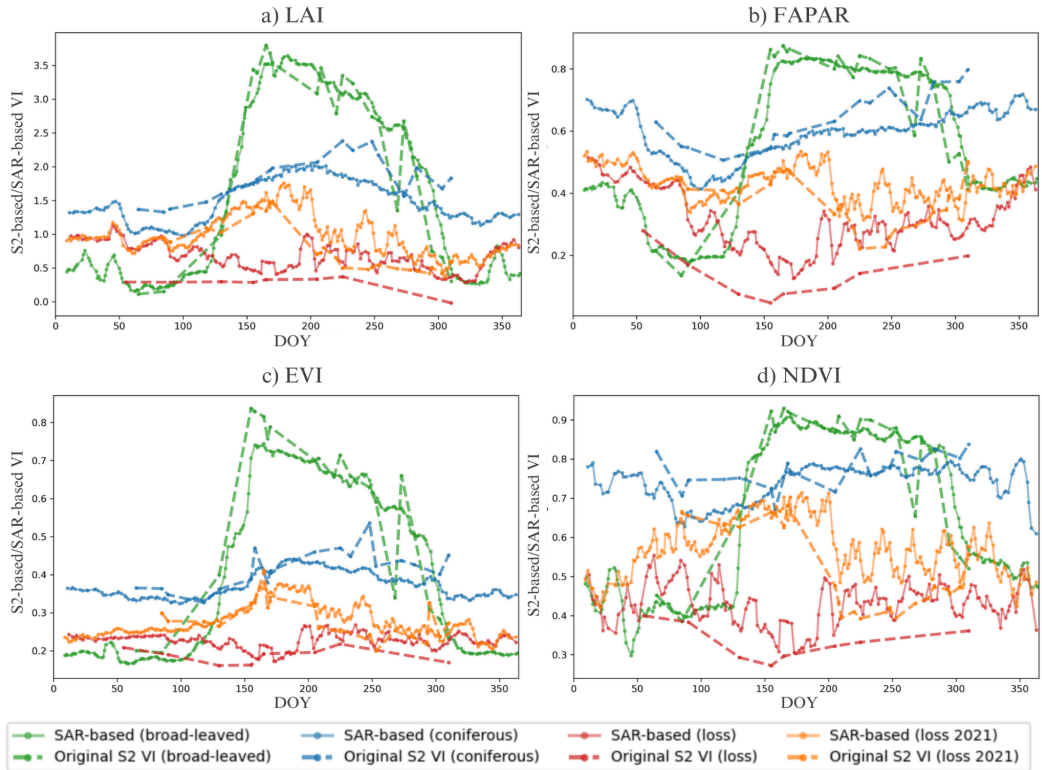

Fig. 9. TS of S2-based and estimated (SAR-based) VIs for LAI (a), FAPAR (b), EVI (c) and NDVI (d).

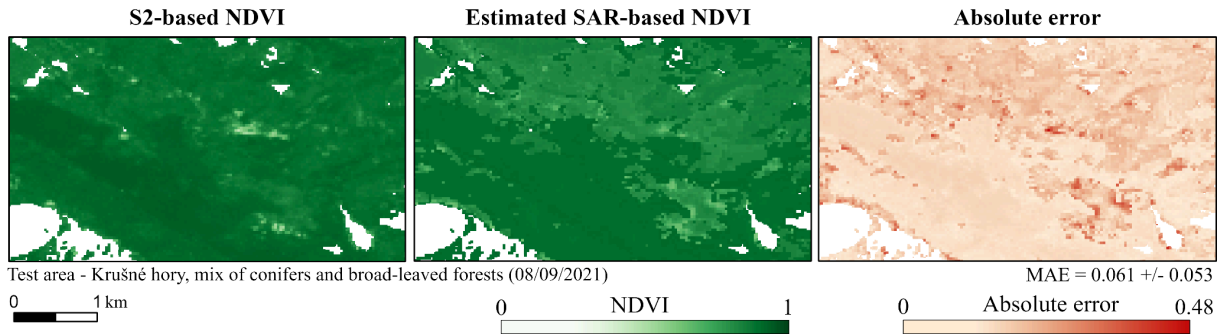

Fig. 10. Comparison of S2-based NDVI (a), estimated SAR-based NDVI (b) and absolute errors (c) for the entire Krušné hory test area.

# Discussion

In this study, **a successful estimation of four optical VI TS** was presented using SAR and ancillary data with the use of ML techniques for forest monitoring. For this purpose, a multi-modal TS dataset generated using the MMTS-GEE tool, including spatially and temporally aligned S1, S2, weather- and DEM-based features for 2021 over Czechia. The emphasis was placed on a successful estimation of VIs for both healthy and disturbed forest areas by selecting an appropriate input dataset that contains TS of both healthy and disturbed forest areas. As expected, when using only healthy datasets, the inference on healthy case studies worked well with very high accuracy, but the estimations on disturbed case studies showed a TS corresponding to a healthy forest area, i.e., that model was not able to detect the forest change. Furthermore, the permutation feature importance revealed that in these cases, the SAR and other satellite-based features had a negligible impact on the final results, while the information about time and forest type had a crucial effect (Fig. 8). This means that only information on time and forest type would be sufficient to estimate the NDVI TS for healthy forests. On the other hand, when disturbed data were included, TS estimations for disturbed case studies showed expected results for both healthy and disturbed forest areas where the VH polarization was the most important SAR feature, being 4th most important feature after forest type and time features. High importance of VH can be most likely due to the its high correlation with vegetation density and VIs, as also found in previous studies (Alvarez-Mozos et al. 2021; Filgueiras et al. 2019; Holtgrave et al. 2020; Jiao, McNairn, and Dingle Robertson 2021).

**TS analysis** showed good agreement between S2- and SAR-based VIs with clear differentiation between broad-leaved and coniferous forests, and the ability to estimate the timing of disturbance events. The SAR-based VIs demonstrated consistent seasonal patterns, though some overestimation was observed, particularly during winter. Additionally, fluctuations in the TS for disturbed areas were noted, indicating variability in forest conditions before and after disturbance. It can be caused by several factors, e.g., higher heterogeneity and higher soil roughness of clear-cut areas with possible left-overs and/or disturbed soil after the use of heavy machinery; limited attenuation of SAR signal by tree crowns causing high contribution of soil with greater influence of environmental factors (e.g., precipitation and increased soil moisture from bare land, snow cover or frozen conditions). The noisy behavior of estimated SAR-based VIs needs to be improved to achieve more timely estimations of disturbance events by e.g. adding more disturbed forest areas, ideally also disturbed broad-leaved forests, to the training dataset. The lower VI values for case study 3 during the pre-disturbance period observed in this study indicate a less than perfect condition of forests, suggesting that it is possible to monitor the health status of forests using both the S2-based and the estimated SAR-based VI, especially using NDVI and FAPAR. This finding can serve as an important basis for studying forest health or forest degradation, such as the stages of bark beetle infestation, using SAR TS.

**The detection of forest changes** using each SAR-based VIs was considerably more timely than for the original S2-based VI TS. On the selected case study areas, the SAR-based VIs were able to achieve a sub-weekly accuracy, specifically up to 4 days in detecting changes, whereas the S2-based VI exhibited over a month of uncertainty between the last time step of a healthy state of the forest and the first one in a disturbed state. This uncertainty depends on the availability of cloud-free S2 measurements over the study area, while, on the other hand, all possible S1 acquisitions can be used to generate a SAR-based VI. The number of S1 acquisitions in 2021 was greater than 230 in two case studies and around 175 in the other two, while the number of cloud-free S2 measurements for the same areas ranged from 9 to 21. This brings great advantages when monitoring forest changes in applications where near real-time information is required, but also in cases where the exact date of change needs to be determined retrospectively. However, it should be noted that this study was carried out for 2021, where both S1-A and S1-B were operational; therefore, full potential of the proposed methodology can be used for S1 data prior to 2022 or after the planned launch of the S1-C satellite. Moreover, new opportunities will be open for VI estimation using SAR data when freely available L- and P-band data will be available from the NISAR, Rose-L and Biomass missions. Multi-band SAR data fusion could benefit forest monitoring due to the different behavior of L- and C-band signals for various forest types and also for forest degradation. Moreover, SAR polarimetry with quad-polarizations of the new SAR missions would also be beneficial to expand the input feature space.

Due to **S2 data cloud-masking**, in some periods of the year, only a few S2 measurements were available for analysis. Even the state-of-the-art cloud masking algorithm for S2 data, CloudScore+, was used, some defective pixels could still be present in the dataset, causing errors in the input dataset, especially in wintertime. Although it is a potential issue in ML model training, the inclusion of available winter data was crucial for all-year VI estimation. In dos Santos, Da Silva, and do Amaral (2021) and dos Santos et al. (2022), a better accuracy was found in the dry season compared to the rainy season, which can also be attributed to the higher availability of S2-S1 data combinations. Further improvements are needed to reduce errors in periods with a lack of data, e.g. by expanding the time range of the data, such as adding data from more years, not only from 2021. Extending the proposed methodology to other geographic and climatic conditions would be important to assess its accuracy globally in areas with cloudy, less cloudy and no cloudy conditions.

**Spatial analysis** performed well on the selected case study area with an average MAE of 0.06 and a clear differentiation between forest types, while the highest MAE observed was 0.48 in the NDVI estimation. Higher MAEs in spatial analysis usually had two main sources: 1) the inaccuracy of the underlying forest type datasets and 2) errors in areas of current forest changes. As found in the permutation feature importance analysis, forest type is a leading feature due to which it is possible to differentiate between the seasonality of coniferous and broad-leaved forests. Therefore, the accuracy of the forest type dataset in a required spatial resolution is a fundamental element in estimating the VIs both spatially and in TS. Errors due to recent forest changes are related to the fact that a disturbed or cleared forest can have a very high variance in SAR backscatter due to structural changes and various environmental effects, as discussed above. Furthermore, areas of broad-leaved forest loss were not included in the training, so it is not possible to detect broad-leaved forest loss using the proposed methodology.

**The incorporation of ancillary data** consistently enhanced the accuracy of SAR-based VIs. In the proposed pipeline, including DEM-based features alongside SAR-based ones resulted in an average 3% increase in the $R^2$ value. Elevation and slope information can provide information about differences in forests (height, density, tree crown structure) that are specific for different altitudes and slopes. The presented results are similar to those in Roßberg and Schmitt (2022), where models using more input data (SAR, DEM and land cover) achieved better results. A moderate correlation between SAR and DEM features (slope and elevation) were found for coniferous forests in Paluba, Le Saux, et al. (2024). Another DEM-based feature, the LIA holds important information about the differences between acquisition orbits and paths, which have a high correlation with S1 polarizations for both coniferous and deciduous forests (Paluba, Le Saux, et al. 2024). Improvement in the accuracy of NDVI estimation from SAR data by adding LIA to enhance the feature space was also found in Filgueiras et al. (2019). The inclusion of both DEM- and weather-based data improved the $R^2$ by another 1% on average. The influence of accumulated precipitation 12 hours prior to SAR acquisitions on the results can be attributed to the findings that precipitation, which increases soil moisture, affects the backscatter from forests in Benninga, van der Velde, and Su (2019). A correlation was found between SAR polarizations and polarimetric parameters and temperatures (Rüetschi, Small, and Waser 2019; Olesk et al. 2015; Paluba, Le Saux, et al. 2024), while temperatures below 1°C were found to negatively affect SAR backscatter in Benninga, van der Velde, and Su (2019); Ranson and Sun (2000). Moreover, the tested optical VIs TS have a clear relationship with temperature, i.e., low values in wintertime and high values in summer.

As **ML techniques** are nowadays frequently used tools in RS, **AutoML** methods can save time for researchers by automating repetitive and time-consuming tasks, efficiently exploring the algorithm and hyperparameter space and improving the performance by using ensembles of the best found models. Given these advantages, AutoML makes ML accessible to users with less ML experience and allows them to develop competitive models. On the other hand, it has its limitations, e.g., it may not capture domain-specific knowledge as effectively as manual selection of algorithms. It has limited transparency and interpretability in model selection and hyperparameter optimization. There is also a lack of customization because it has a limited support for highly specialized or novel techniques. Moreover, most of the AutoML libraries are prepared for tabular data, but some are for image data, such as Auto-Keras (Jin et al. 2023). The use of AutoML is in general an effective but time- and resource-consuming process of searching for the best pipelines. Therefore, recent studies are focusing on developing the so-called *Green AutoML* approaches (Tornede et al. 2023) allowing one to consider the carbon footprint of the hyperparameter optimization process or proposing methods that are competitive in performance with robust AutoML frameworks, but require only a few seconds to find the best model due to improved meta-learning, for instance, TabPFN (Hollmann et al. 2023).

**ML model comparison** revealed that the two fine-tuned baseline algorithms (RFR and XGB) with AutoML, RFR and XGB slightly outperformed the best ensemble models identified by AutoML, even after 12 hours of optimization. The lowest MAEs for LAI and FAPAR were achieved by RFR, and for EVI and NDVI by XGB. Although the best performing hyperparameter combinations of each model were very similar, it is important to consider the time and computational resources needed to achieve these results. Without doubt, the fastest and the least computational heavy algorithm for training and fine-tuning was the XGB. The XGB was also very effective due to its built-in early stopping mechanism preventing overfitting by stopping further tree growth. The significantly slower fitting times of RFR compared to XGB, in about 30- to 70-times slower, were caused mainly due to the MAE evaluation criteria, also stated in the scikit-learn RFR documentation. Interestingly, during the tests with different optimization lengths, the final ensemble model in auto-sklearn was at some points composed entirely of GB algorithms (after 1 minute for EVI, 10 minutes for FAPAR and NDVI, and 60 minutes for LAI). Extending the AutoML optimization time beyond these lengths led to either negligible performance improvements (less than 0.25% in R²) or even a decrease in performance, particularly for EVI. This also demonstrates the effectiveness of GB and its advanced variant, the XGB, as the state-of-the-art algorithm among traditional ML methods (Grinsztajn, Oyallon, and Varoquaux 2022), for estimating VIs for forests using SAR and other ancillary data.

**In general**, the proposed methodology proved to be appropriate for estimating standard optical VIs with high accuracy, even with computationally effective and fast ML algorithms, such as XGB. Compared to optical VIs, which are prone to atmospheric effects, the SAR-based VI estimations can provide all-year coverage with up to 237 measurements / year using two Sentinel-1 satellites. SAR-based VIs can serve as an alternative source of information on forest seasonality and dynamics and can benefit from the timely monitoring of forest health status and changes. Moreover, in comparison to global or regional VI products, providing a good temporal resolution (such as 10 days in CGLMS LAI and FAPAR), and better spatial resolution (20 m in a 10 m grid) can be achieved using a SAR-based VI.

# Conclusion

This study successfully demonstrated the feasibility of using SAR data and ML techniques to estimate well-known and widely used optical VIs, including NDVI, EVI, LAI and FAPAR, for monitoring temperate forests in Czechia, within a Central European context. Cloud-independent SAR-based VIs proved to be a reliable alternative to optical VIs with high accuracies ($R^2$ between 70-86%) and low errors (0.055-0.29 of MAE). Traditional ML algorithms, specifically RFR and XGB slightly outperformed AutoML in VI estimation, while XGB stood out as the most efficient and computationally lightweight option, offering fast training times and effective prevention of overfitting. Including DEM- and weather-based ancillary features improved the model performance, proving their importance for VI estimation alongside SAR data, with VH polarization identified as the most important SAR feature. The results highlight the potential of SAR-based VI estimations, which provide year-round coverage and can offer higher spatial and temporal resolution compared to traditional optical VIs, achieving up to 240 measurements / year when both Sentinel-1 satellites are operational. This makes SAR-based VIs a valuable tool for monitoring forest seasonality, health, and dynamics, and for detecting disturbances more quickly and accurately than S2-based VIs, enabling sub-weekly monitoring of forest changes. Future research should extend this methodology to other geographic regions and integrate data from upcoming SAR missions, which could further enhance the accuracy and applicability of SAR-based forest monitoring.

# Acknowledgments

This work was supported by the Charles University Grant Agency – Grantová Agentura Univerzity Karlovy (GAUK) Grant No. 412722 and the European Union's Caroline Herschel Framework Partnership Agreement on Copernicus User Uptake under grant agreement No. FPA 275/G/GRO/COPE/17/10042, project FPCUP (Framework Partnership Agreement on Copernicus User Uptake). The authors would like to thank the Spatial Data Analyst project (NPO_UK_MSMT-16602/2022) funded by the European Union – NextGenerationEU, for providing computational resources needed for AutoML. Daniel Paluba would like to thank the Erasmus+ programme for the financial support during his research stay at the Φ-lab, European Space Agency (ESA) in Frascati, Italy.

# Disclosure statement

No potential conflict of interest was reported by the author(s).

# Data Availability Statement

Open-access data in the Google Earth Engine platform were used in this study. The *Copernicus High Resolution Layer Forest Type* layers were accessed from land.copernicus.eu/en/products/high-resolution-layer-forest-type. Data preprocessing pipelines in GEE to generate the multi-modal time series dataset (MMTS-GEE) is available on GitHub at github.com/palubad/MMTS-GEE. The best found RFR, XBG and Auto-sklearn models, and all input datasets (described in sections *Training and testing data validation*, *Case studies for healthy and disturbed forest time series analysis* and *Test area for spatial comparison of results*) are available on GitHub, at github.com/palubad/SAR-based-VIs. [**Note: During the review process, the GitHub repository is closed.**]

# Appendices

Appendix I. Auto-sklearn regressors, feature and data preprocessors with the number of their hyperparameters (# λ). Based on the source-code of auto-sklearn on GitHub (Feurer et al. 2023) [accessed on 26.09.2023].

| Regressor | # λ | Feature preprocessor | # λ | Data preprocessor | # λ |
|---|---|---|---|---|---|
| adaboost | 4 | densifier | - | Cat. encoder | 3 |
| ard_regression | 8 | Extra Trees Regr. Prep. | 9 | imputation | 2 |
| decision_tree | 8 | fast ICA | 4 | Category Shift | - |
| extra_trees | 9 | feature agglomeration | 4 | rescaling | 10 |
| gaussian_process | 3 | kernel PCA | 5 | Cat. minority coalescer | 2 |
| gradient_boosting | 12 | Random Kitchen Sinks | 2 | | |
| k_nearest_neighbors | 3 | linear SVM prepr. | 3 | | |
| liblinear_svr | 4 | no preprocessing | - | | |
| libsvm_svr | 7 | Nystroem kernel approx. | 5 | | |
| mlp | 15 | PCA | 2 | | |
| random_forest | 9 | Polynomial Features | 3 | | |
| sgd | 11 | Random Trees Embedding | 7 | | |
| | | select perc. Regression | 2 | | |
| | | Univariate Feature Selection based on rates | 3 | | |
| | | truncatedSVD | 1 | | |

Appendix II. Fine-tuning results for RFR and XGB for LAI (*a* and *b*), for FAPAR (*c* and *d*) and for EVI (*e* and *f*). The top 20 results with the lowest MAE are highlighted with yellow, the red highlighted cell represents the best performing hyperparameter combination.

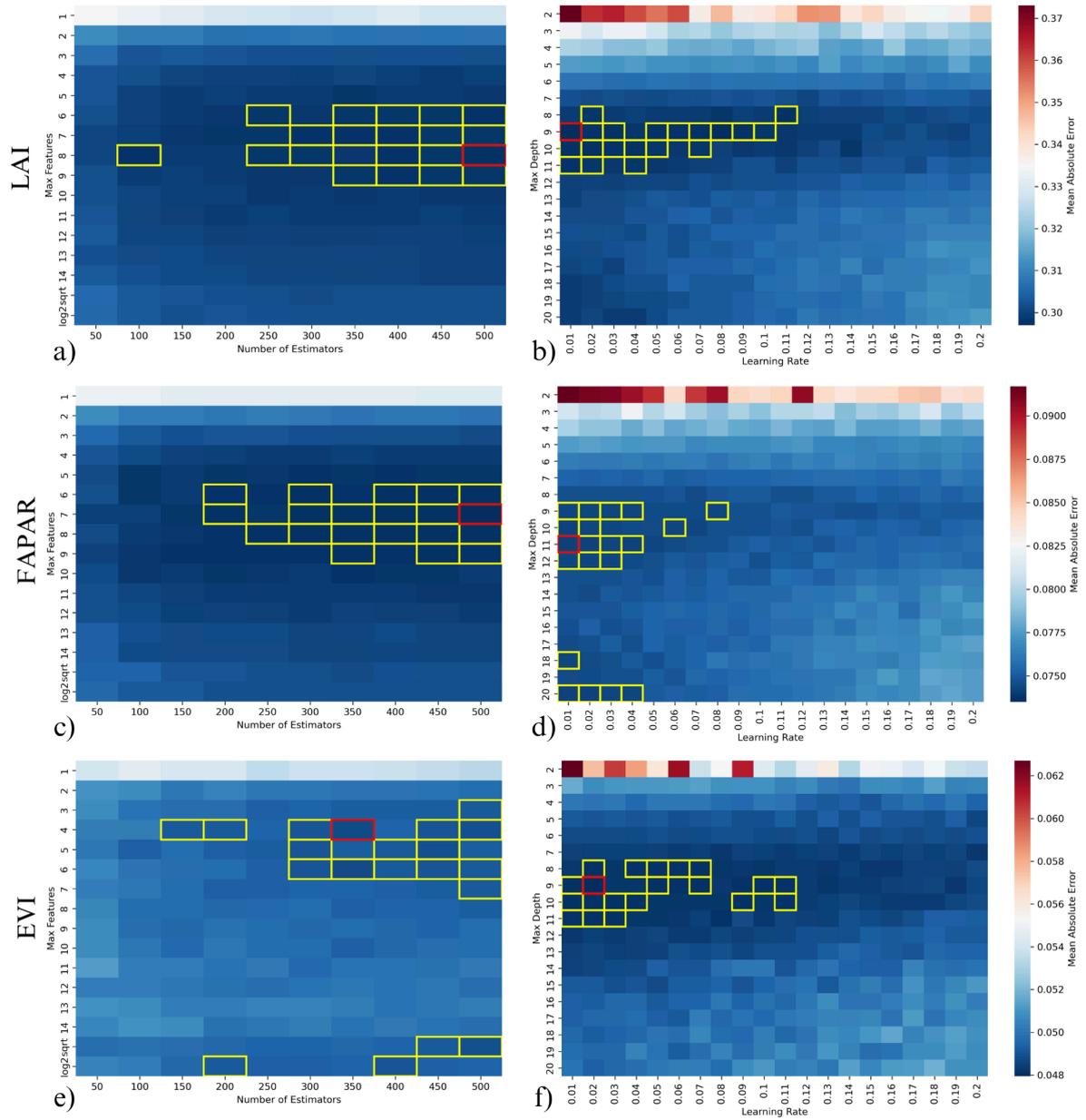